\documentclass[dvipsnames,sigconf=true, nonacm=true, review=false, anonymous = false,]{acmart}
\usepackage{graphicx} 
\usepackage{url}
\usepackage{xcolor}
\usepackage{amsmath}

\copyrightyear{2026}
\acmYear{2026}
\setcopyright{cc}
\setcctype{by}
\acmConference[GECCO '26]{Genetic and Evolutionary Computation Conference}{July 13--17, 2026}{San Jose, Costa Rica}
\acmBooktitle{Genetic and Evolutionary Computation Conference (GECCO '26), July 13--17, 2026, San Jose, Costa Rica}
\acmDOI{10.1145/3795095.3805129}
\acmISBN{979-8-4007-2487-9/2026/07}

\ccsdesc[500]{Theory of computation~Design and analysis of algorithms}
\ccsdesc[500]{Theory of computation~Bio-inspired optimization}

\keywords{Benchmarking, Multi-Objective Optimization, Transformations}

\title[Pareto-preserving Search Space Transformations in MOO]{Exploration of Pareto-preserving Search Space Transformations in Multi-objective Test Functions} 

\author{Diederick Vermetten}
\orcid{0000-0003-3040-7162}
\affiliation{
  \institution{Sorbonne Universit\'e, CNRS, LIP6}
  \city{Paris}
  \country{France}}
\email{diederick.vermetten@lip6.fr}

\author{Jeroen Rook}
\orcid{0000-0002-3921-0107}
\affiliation{
  \institution{Paderborn University}
  \city{Paderborn}
  \country{Germany}}
\email{jeroen.rook@uni-paderborn.de}

\begin{document}

\begin{abstract}

Benchmark problems are an important tool for gaining understanding of optimization algorithms. Since algorithms often aim to perform well on benchmarks, biases in benchmark design provide misleading insights. In single-objective optimization, for example, many problems used to have their optimum in the center of the search domain. To remedy these issues, search space transformations have been widely adopted by benchmark suites, preventing algorithms from exploiting unintended structure. 

In multi-objective optimization, problem design has focused primarily on the objective space structure. While this focus addresses important aspects of the multi-objective nature of the problems, the search space structures of these problems have received comparatively limited attention. In this work, we re-emphasize the importance of transformations in the search space, and address the challenges inherent in adding transformations to boundary-constraints problems without impacting the structure of the objective space. We utilized two parameterized, bijective transformations to create different instantiations of popular benchmark problems, and show how these changes impact the performance of various multi-objective optimization algorithms. In addition to the search space transformations, we show that such parameterized transformations can also be applied to the objective space, and compare their respective performance impacts.

\end{abstract}

\maketitle

\section{Introduction}

In the area of black-box optimization, benchmark problems play a major role. Not only do they offer a way to compare algorithms on a set of standardized optimization scenarios, but their design generally aims to provide insight into the strengths and weaknesses of optimizers. This can be achieved by focusing on specific types of optimization challenges in different problems of a benchmark suite, for example in COCO's popular BBOB suites~\cite{hansen2021coco}. However, benchmark design does not only influence algorithm comparison, but also implicitly guides algorithm development, as shown by an inherent desire to improve performance on the commonly used problem sets.

Given the broad influence of benchmark design, it is important to carefully consider how decisions made in this process might impact algorithm design. For example, in single-objective optimization, benchmark problems with their optimum located in the center of the search space can lead to biases in algorithm comparison. In this case, algorithms which initialize around the center (or in the center exactly), can be unfairly seen as better performing, independent of their actual search strengths~\cite{kudela2022critical}. A popular way to remedy this type of issue is to use different instances of a given base problem, which makes it easier to identify when an algorithm is exploiting unintended properties of a function~\cite{hansen2009real}. 

When moving from single to multi-objective optimization (MOO), an additional challenge is introduced in the form of the structure of the objective space. Where many single-objective algorithms are invariant to monotonic transformations of the objective space, this is no longer the case on the multi-objective domain. As a result of this added challenge, it is natural to focus on different structures of the objective space to understand how optimizers handle them. Many of the popular benchmark suites for multi-objective optimization have thus been designed with a major focus on objective space structures, with differences between their included problems primarily being used to compare the ability of optimizers to handle these challenges in the objective space~\cite{deb1999mopdef,huband2006WFG}. 

While the focus of benchmark design on the objective space has led to useful progress in algorithm design, this focus seems to have distracted from the investigation of search space structures. However, the challenges in single-objective optimization do not simply disappear when the problems become multi-objective. Thus, we argue that it remains important to investigate the ability of optimizers to handle different types of search space structures, and to investigate whether existing benchmark sets unintentionally favor biased optimizers by the nature of their design focus.

In this paper, we explore how two bijective search space transformations can be applied to existing multi-objective benchmark problems, and how this impacts the behavior of several commonly used optimizers. The first transformation is a component-wise Beta cumulative distribution function, modifying volumetric distributions of the search space. The second transformation rotates the search space, altering variable-objective alignment without changing problem topology. Both transformations preserve the reachability and structure of the objective space, they offer a useful methodology for verifying the stability of optimizers to changes in the search space.

\section{Related Work}\label{sec:rel_work}

Zitlzer et al.~\cite{Zitler2000ZDT} proposed the ZDT benchmark suite that consists of 6 bi-objective test functions. These problems were intended to assess the ability of multi-objective optimizers to handle specific optimization challenges. To achieve this, each function provides a specific problem characteristic, such as convexity, nonconvexity, discreteness, multimodality, and nonuniformity. 
Except for nonuniformity, all other features focus on the objective space properties of the function. 
The DTLZ benchmark suite~\cite{Deb2005DTLZ} holds 9 test problems that are, in relation to ZDT, scalable for both the number of decision variables and objectives. The DTLZ problems again focus on having different objective space characteristics. Huband et al.~\cite{huband2006WFG} provide a critical review on the quality of existing benchmark suites at the time, including ZDT and DTLZ. They conclude that these test functions lack certain challenges, such as being deceptive (i.e. multi-global landscapes\cite{preussMultimodalOptimizationMeans2015}) or non-separablilty. To address these shortcomings, they proposed the WFG toolkit, which contains 9 scalable test problems, but also includes transformation functions to vary the complexity of the objective space. These three benchmark suites are still among the most widely-used benchmark suites in MOEA research papers.

Another source for benchmarking problems in MOO has been the CEC competitions. Usually, these competitions focus on a specific theme, such as multi-modality in 2019~\cite{yue2019MMF}. Contrary to the ZDT, DTLZ, and WFG these benchmark sets are usually not composed to be diverse in their problem characteristics, but are rather comprised with problems of increasing difficulty. Over the years, many other benchmark suites have been proposed, where the majority focused on varying Pareto shapes~\cite{masuda2016,kenny2025multi}, or problems suitable for many objectives~\cite{cheng2017MaF,liMultilineDistanceMinimization2018}.

Orthogonal to (scalable) benchmarking suites, problem generators have been proposed. The central challenge with these generators is that ideally the problem characteristics, Pareto sets and Pareto fronts are known. One example of a generator that meets all these requirements is the multiobjective multiple peaks models~\cite{Schapermeier2023apeekaboo}.

In single-objective optimization (SOO), transformations play a central role in the creation of benchmark suites. In particular, COCO's BBOB benchmark suite~\cite{hansen2021coco} applies transformations (instances) to base functions to control problem characteristics. In contrast to MOO, transformations for SOO tend occur in the search space, such as translations and rotations, as they change how algorithms explore and exploit the optimization landscape while preserving the underlying problem function. Conversely, transformations of the objective space are typically of less importance since both landscape features and algorithmic operators tend to be comparison-based and thus invariant to e.g. scaling of objective values~\cite{yin2024impact}. Without search space transformations, benchmarking results may be influenced by structural biases in either the problems or the algorithms~\cite{kudela2022critical, vermetten2022bias}, for example when optima are located at the origin, which can advantage center-biased algorithms. Unfortunately, in MOO, search space transformations generally interact with dominance relations and the geometry of the Pareto set, making it not trivial to directly transfer transformations seen in SOO.

Several studies have investigated search space transformations in the multi-objective setting, with a particular focus on rotations~\cite{iorioRotatedTestProblems2006}.
However, such transformations can introduce boundary-related artifacts, as rotated decision variables may interact with bounded domains in a way that alters the feasible Pareto set and, consequently, the shape or extent of the Pareto front. Avoiding these effects typically requires substantial modifications to the problem formulation~\cite{igel2007covariance}. The bi-objective BBOB test suite~\cite{Brockhoff2022biobjbbob}, applies search space transformations to individual functions before they are combined into a bi-objective optimisation problem, which may lead to changes in objective space characteristics, including distortions of the Pareto front geometry~\cite{Schapermeier2023apeekaboo}. 

\section{Instance Transformations}\label{sec:methods}
To analyze search space transformations in a multi-objective context, we have to carefully consider the original problems we utilize. Specifically, we want to ensure that no regions of the objective space become unreachable, or that new points get introduced which might dominate existing Pareto-optimal solutions. This fact, combined with the box-constrained nature of the considered problems, means we can not use the standard transformations (translation, rotation, scaling). Instead, we use a modified rotation function which ensures we have a bijection $[0,1]^n\rightarrow[0,1]^n$, as well as a domain warping methodology based on the Beta-CDF with the same bijective property. 

Because these transformations are bijective, we ensure that the shape of the problem in the objective space remains consistent. However, this does not mean the volume distribution between non-dominated sets remains consistent. Because of this, we include a random search method in all our comparisons.

\begin{figure}
        \centering
    \includegraphics[width=0.95\linewidth]{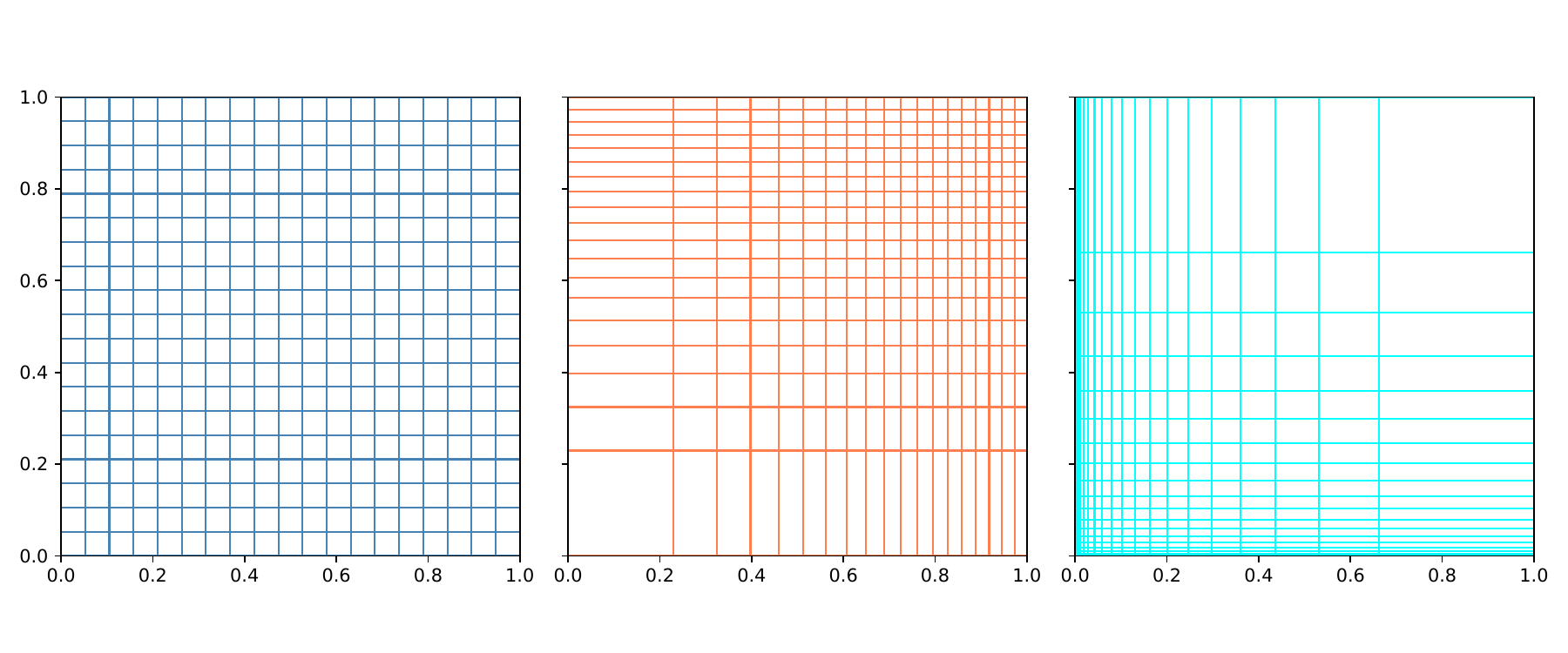}
    \caption{The impact of the Beta-CDF based transformation on the 2-dimensional unit grid, for two select pairs of $(\alpha, \beta)$. Left: original search space. Middle: Beta-CDF transformed with parameters $(0.5,1.0)$. Right: Beta-CDF transformed with parameters $(2.0,0.5)$.}
    \label{fig:transform_ab}
\end{figure}

\subsection{Beta-CDF based transformation}

The first set of search space transformations we consider is based on the cumulative density function of the Beta distribution. This type of transformation has been used in the context of Bayesian optimization to warp the input space when dealing with non-stationary functions~\cite{snoek2014input} or to transfer surrogate models between different problem instances~\cite{pan2025transfer}.  
Specifically, for the transformation we use, we have:
\begin{align*}
    x_i'&=\textit{BetaCDF}(x_i, \alpha, \beta)\text{,\ } \forall x_i\\
\end{align*}
Where $\alpha$ and $\beta$ are the parameters for the Beta distribution. This transformation is thus a coordinate-wise bijection and can be inverted by using the percent point function of the same distribution. Even though the analytical form of the Beta-CDF is not available when the shape parameters are not integers, we can make use of an approximation as implemented in the \textit{scipy} package~\cite{virtanen2020scipy}.   

Note that to avoid having to vary too many parameters, we opt to keep $\alpha$ and $\beta$ constant between dimensions. To illustrate the impact the Beta-CDF transformation has on a 2-dimensional search space, we show two instantiations of this transformation in Figure~\ref{fig:transform_ab}. As we transform the space coordinate-wise, we do not introduce any new dependencies between variables, and thus preserve any potential axis-aligned structures from the original space. However, since the density of the different areas changes significantly, this transformation can still impact the ease of finding these optimal regions. 

In addition to the transformation of the search space, we can also use this same methodology on the objective space. Similar to previous work, which applied transformations on the objectives to identify the impact of L-shaped fronts on algorithm behavior~\cite{kenny2025multi}, we can use the Beta-CDF to compare the relative effects of changes to search and objective spaces. Note, however, that for this transformation we require a bounded space, so the transformation is only applied to the $[0,1]^d$ region of the original objective space.

\subsection{Sphered rotation transformation}
The second transformation method we employ is based on rotation. As discussed in Section~\ref{sec:rel_work}, the impact of rotation in multi-objective optimization benchmarking has been previously studied and shown to be quite impactful to optimizer performance~\cite{iorioRotatedTestProblems2006}. However, since rotation in a box-constrained domain is not bijective, these analyses have required a modification of the problem functions themselves. Since we aim to utilize problem-independent transformation procedures, we make a slight modification to the standard rotation procedure. We employ an intermediate mapping to the unit-hypersphere, which is then rotated before being mapped back to the unit-hypercube. Specifically, this transformation with a given rotation matrix $R$ works as follows:
\begin{align*}
    \boldsymbol{z} &= 2\boldsymbol{x}-1 \tag*{\text{(position center at origin)}}\\ 
    \boldsymbol{y}  &= \frac{\boldsymbol{z}}{ \left\| \boldsymbol{z} \right \|_{\infty}} \tag*{\text{(project to hypersphere)}}\\
    \boldsymbol{y^R} &= R\boldsymbol{y} \tag*{\text{(rotate)}}\\
    \boldsymbol{z'} &= \boldsymbol{y}^R\ \left \| \boldsymbol{y}^R \right\|_{\infty} \tag*{\text{(reverse projection)}}\\
    \boldsymbol{x'} &= \frac{\boldsymbol{z'}+1}{2} \tag*{\text{(reverse position)}}
\end{align*}

\begin{figure}
        \centering
    \includegraphics[width=0.95\linewidth]{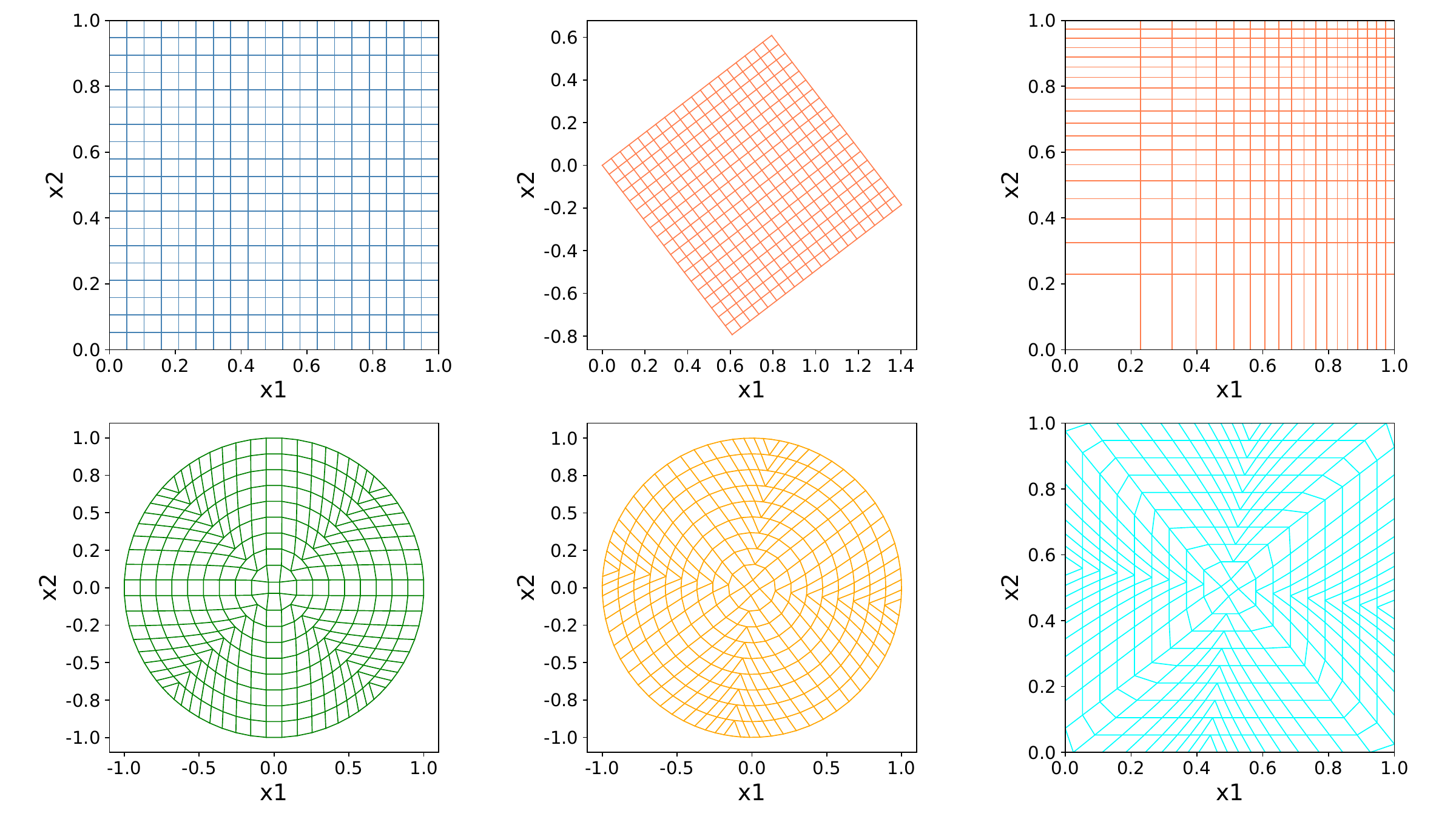}
    \caption{The different steps of the rotation-based transformation. Top row: starting grid, standard rotation (after centering) and final result when scaled back to the original domain. Bottom row: steps to create the sphered rotation. First, the grid is mapped to the unit sphere, then rotation is applied, and then the sphere is mapped back to the box and rescaled to the original domain. }
    \label{fig:transform_rotation}
\end{figure}

Figure~\ref{fig:transform_rotation} illustrates this procedure in 2-dimensional space, as well as the problems with the default rotation, where the bijection is not present when the rotation angle is not a multiple of $\frac{\pi}{2}$. Different from the Beta-CDF, rotation does induce dependencies between variables, reducing any axis-aligned structure, such as those in most of the ZDT and DTLZ problems. Throughout this paper, we will refer to this transformation as \textit{sphered rotation}. 

\subsection{Transformation Density Changes}

\begin{figure}
    \centering
    \includegraphics[width=0.95\linewidth]{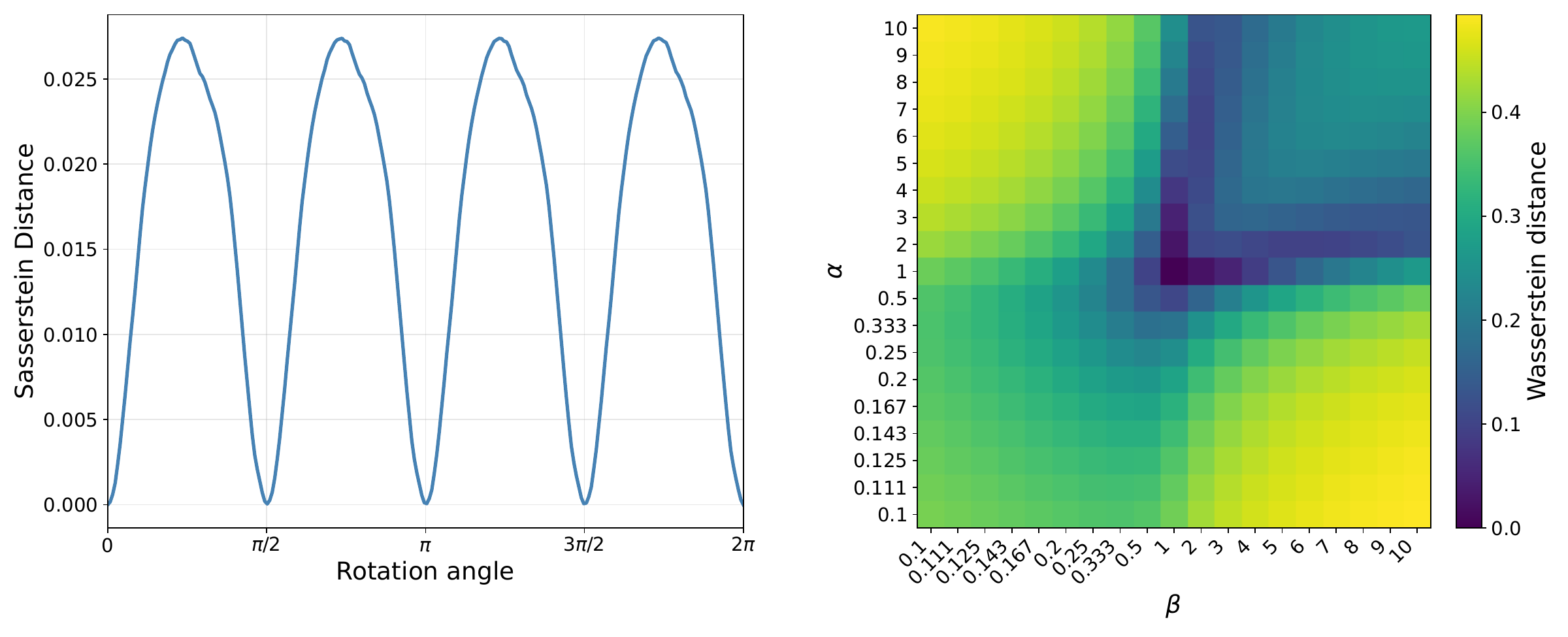}
    \caption{Wasserstein distance between pairwise distances of $500$ random points before and after applying the specified transformations. Left: Sphered rotation in 2-dimensional space, with angle of rotation between $0$ and $2\pi$. Right: AB-transform with $\alpha$ and $\beta$ varied (from $1$ to $10$, with their inverse included)}. 
    \label{fig:density}
\end{figure}

To illustrate how impactful these transformation are, we look at how the pairwise distances between randomly sampled points differ after applying each transformation. Specifically, we look at the Wasserstein distance between pairwise distances of 500 2-dimensional points sampled in the original space. Figure~\ref{fig:density} shows how these distributions differ for different instantiations of the proposed transformations. As we can see, the impact of the sphered rotation is significantly lower than the Beta-CDF based transformation, but still present, except when the angles are multiples of $\frac{\pi}{2}$. 

\section{Experimental Setup}

To analyze the impact of the transformations on the behavior of multi-objective optimization algorithms, we perform a benchmarking study using a set of well-known problem suites. In particular, we employ the problems from the ZDT~\cite{Zitler2000ZDT} and DTLZ~\cite{Deb2005DTLZ} suites, as well as a selection of problems from the MMF generator~\cite{yue2019MMF}. For each of these problem suites, we use the bi-objective versions. For ZDT and DTLZ, we vary the dimensionality of the search space to 2 and 10, while for MMF we can only use the default 2-dimensional versions of the problems except MMF14 and MMF15 (and their `a' variants). Throughout the results, when we refer to dimensionality, we thus refer to the search space dimensionality. 

To create the different problem instances, we employ a set of different parameterizations of the two transformation methods discussed in Section~\ref{sec:methods}. For the BetaCDF-based transformations, we use $\alpha, \beta \in \{0.2,0.5,1.0,2.0,5.0\}^2$, for a total of 25 parameterizations (including the identity at $(1.0,1.0)$). For the sphered rotation, we use four different randomly generated rotation matrices (drawn from the special orthogonal group $SO(N)$), as well as the identity matrix for easy comparison. 

The three algorithms we use are NSGA2~\cite{deb2002fast}, SMSEMOA~\cite{beume2007sms}, and MOEAD~\cite{zhang2007moea}, with population sizes of $10$ and $100$ (using their implementation in PyMOO~\cite{blank2020pymoo}). For MOEAD, a reference vector set, proportional to the population size, is created using the Riesz-Energy algorithm~\cite{falcon2020RieszSenergy}, following PyMOO defaults. We also include a random search baseline for comparison purposes. We perform $10$ independent runs of each algorithm on each problem instance, with each run having a budget of $5000$ function evaluations. 

\begin{figure*}
    \centering
    \includegraphics[width=0.95\linewidth]{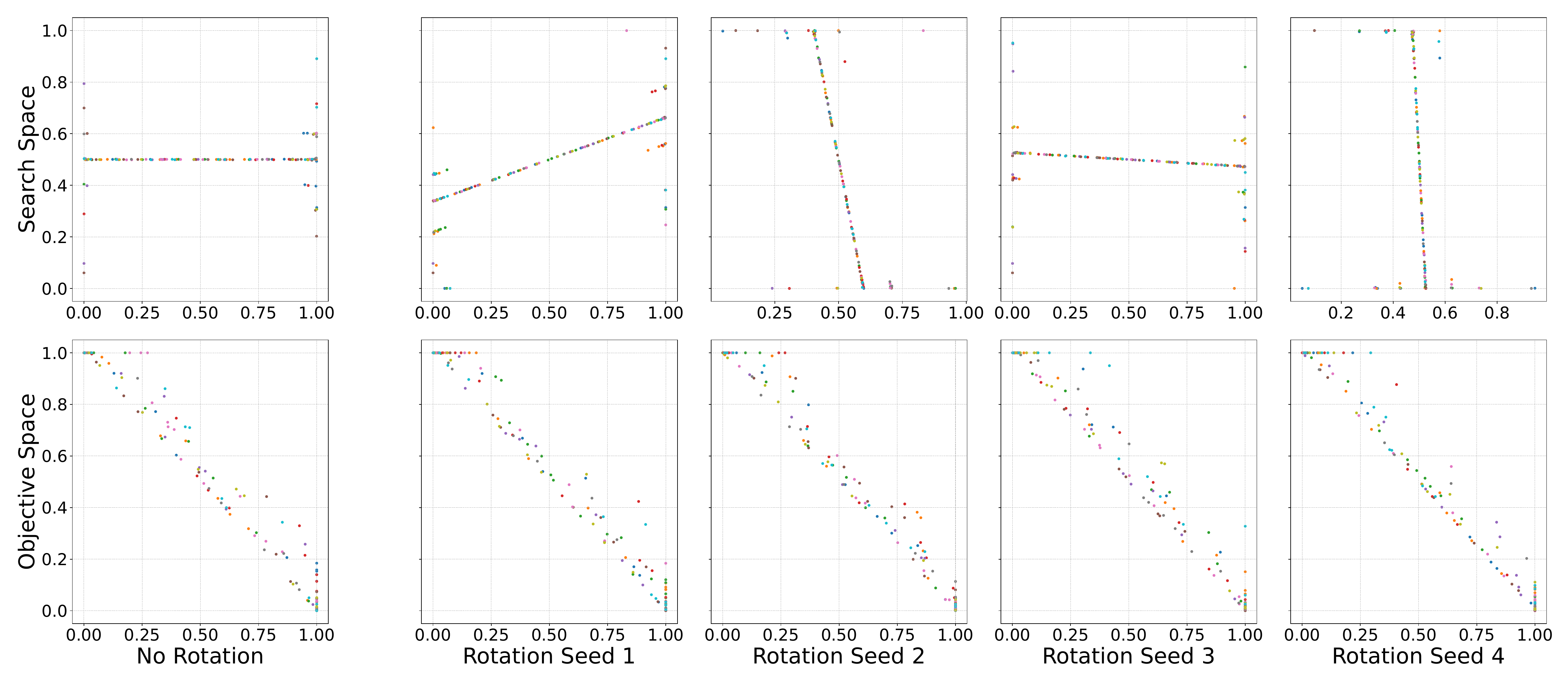}
    \caption{Scatter plots of the non-dominated solutions found by RandomSearch on 2-dimensional DTLZ1 for 5 different instantiations of the sphered rotation (where the first column corresponds to the non-transformed version). First row: search space as seen by the algorithm. 
    Second row: normalized objective space. Each color corresponds to one of 10 independent repetitions.}
    \label{fig:rot_space_example}
\end{figure*}
\begin{figure*}
    \centering
    \includegraphics[width=0.95\linewidth]{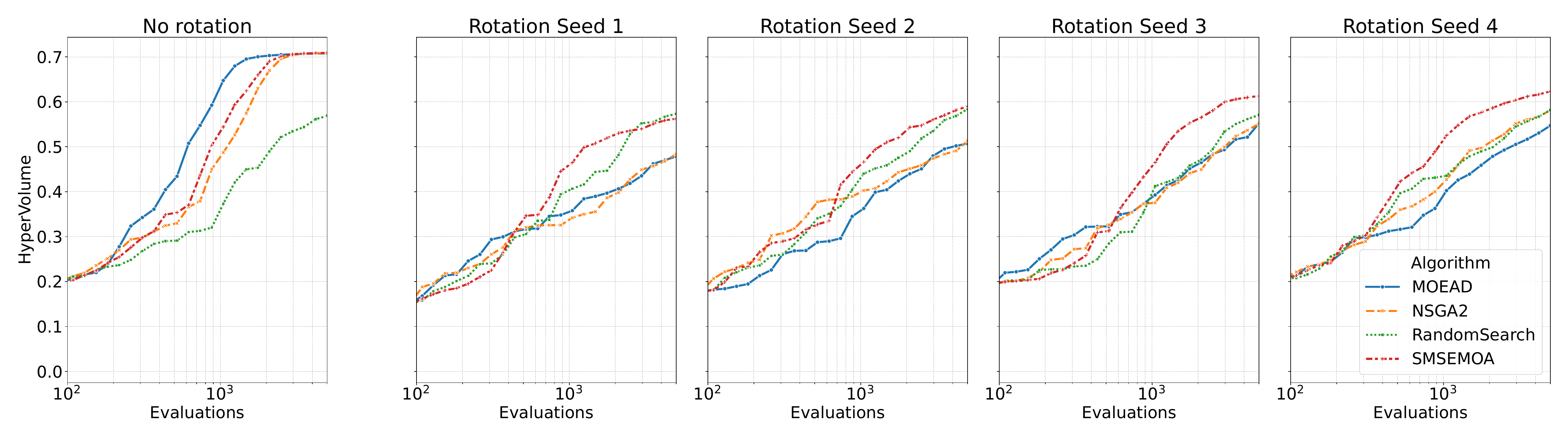}
    \caption{Normalized hypervolume over time (calculated over the unbounded archive) for the 4 used algorithms (population size 100) on 5 instantiations of the sphered rotations on 2-dimensional DTLZ1, where the left-most plot corresponds to the non-transformed version. }
    \label{fig:rot_space_hv_time}
\end{figure*}

To record algorithm performance, we make use of IOHexperimenter for logging~\cite{de2024iohexperimenter}, such that we can utilize a variety of performance perspectives in our analysis, using the IOHinspector package~\cite{vermetten2025mo}. We normalize the objective spaces based on the extrema over all non-dominated solutions across all runs on each function. Depending on the type of analysis, we show either the hypervolume of all non-dominated points sampled during the run (based on an unbounded archive), which allows to visualize performance over time, or only the solutions in the final population of the algorithm. For additional figures, as well as the full reproducibility instructions, we refer to our Zenodo repository~\cite{zenodo}.

\section{Results}

\subsection{Modified rotation procedure}

To understand the impact of the proposed transformations, we first look at how they impact the results of random search. Looking at the non-dominated solutions found by random sampling under different instantiations of the transformation allows us to assess how the density of Pareto-optimal regions has been impacted. For the sphered rotation, the changes in terms of density will occur as a result of the intermediate step to the unit sphere. As such, the angle of rotation determines the expected magnitude the resulting changes, with right-angled rotations leading to zero expected change in the objective space densities, matching the observations from Figure~\ref{fig:density}. 

To illustrate the impact of the sphered rotation on the solutions found by random search, we show in Figure~\ref{fig:rot_space_example} how four different instantiations of this transformation impact both the search- and objective space location of the final non-dominated solutions found across 10 runs of random search. For this figure, we use the 2-dimensional DTLZ1 as the base function, shown as the left-most subplot. From the figure, we can see that the structures in the transformed space indeed have been rotated relative to the original, distorting the axis-aligned nature of the original function. However, we also note that with random sampling, the sphered rotation seems to have a relatively minor impact on the performance of the found non-dominated sets, suggesting that the overall structure of the space near the Pareto-optimal region is mostly preserved.

Next, we look at how the sphered rotation of the search space impacts algorithm performance. To investigate this question, we consider the evolution of hypervolume over time (calculated on an unbounded archive) for the selected multi-objective optimization algorithms. The results for 4 instantiations of the transformation on 2-dimensional DTLZ1 are shown in Figure~\ref{fig:rot_space_hv_time}, where the left-most plot again indicates the original problem. From this figure, we again see the stability of random search performance by looking at the green curve, which remains stable across transformations. However, the other algorithms seem to be rather significantly impacted by the changes to the search space. It is particularly notable that MOEAD, which performed extremely well on the original problem, ends up with worse performance than random search on all four instantiations of the transformation (based on the unbounded archive).  

While not a traditional rotation, the sphered rotation still highlights a potential bias in the design of the optimization algorithms, which exploits the axis-aligned Pareto fronts by using similarly axis-aligned mutation operators. In the implementation we used, all three of these optimizers make use of a polynomial mutation mechanism, which mutates each dimension with a set probability, making changes among axis directions more likely~\cite{iorioRotatedTestProblems2006}.

\subsection{Beta-CDF based transformation}
\begin{figure}
    \centering
    \includegraphics[width=0.99\linewidth,trim=75 20 0 50,clip]{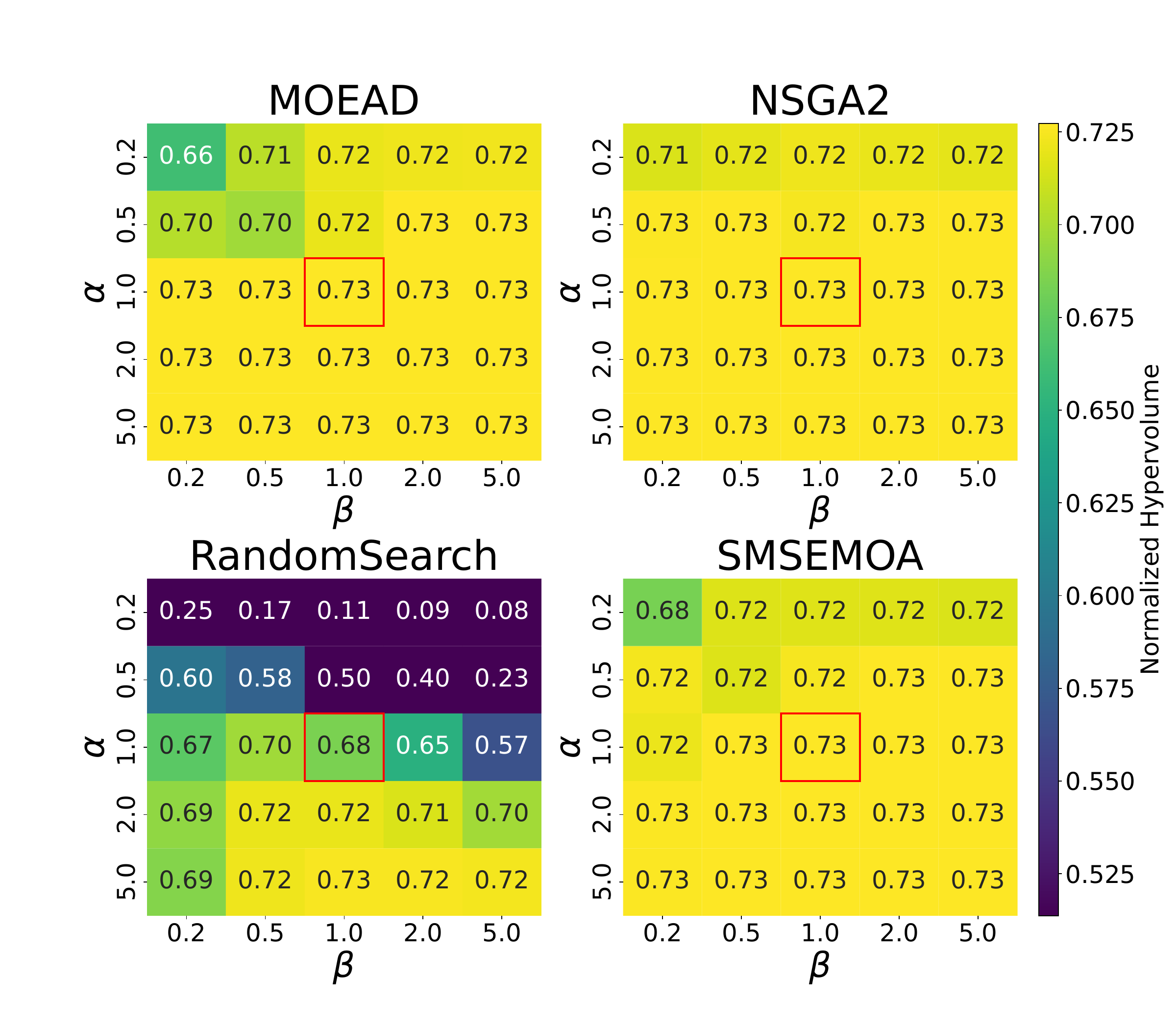}
    \caption{Final hypervolume of the unbounded archive from each of the used algorithms on different instantiations of the BetaCDF transformation (parameterized by $\alpha$ and $\beta$ in the rows and columns respectively). Based on the 2-dimensional ZDT3 problem, using population size 100.}
    \label{fig:final_hv_ab}
\end{figure}

After the sphered rotation, we now look at the impacts of the BetaCDF-based transformation of the search space. As shown in Figure~\ref{fig:transform_ab}, this type of transformation does not change any axis-aligned properties of the space, but rather modifies the density of different regions of the search space. Compared to the sphered rotation, the impact of this type of change is thus much larger for random sampling, as it essentially corresponds to modifying the sampling distribution to be biased to specific regions of the domain. To understand whether this modification introduces new difficulties for the used optimization methods, we can look at the final hypervolume attained on different instantiations of the transformation. Since the BetaCDF is parameterized by two parameters, we sample a grid of $\alpha, \beta \in \{0.2,0.5,1,2,5\}$ to create the different instances of each benchmark problem.

\begin{figure}[!t]
    \centering
    \includegraphics[width=0.95\linewidth]{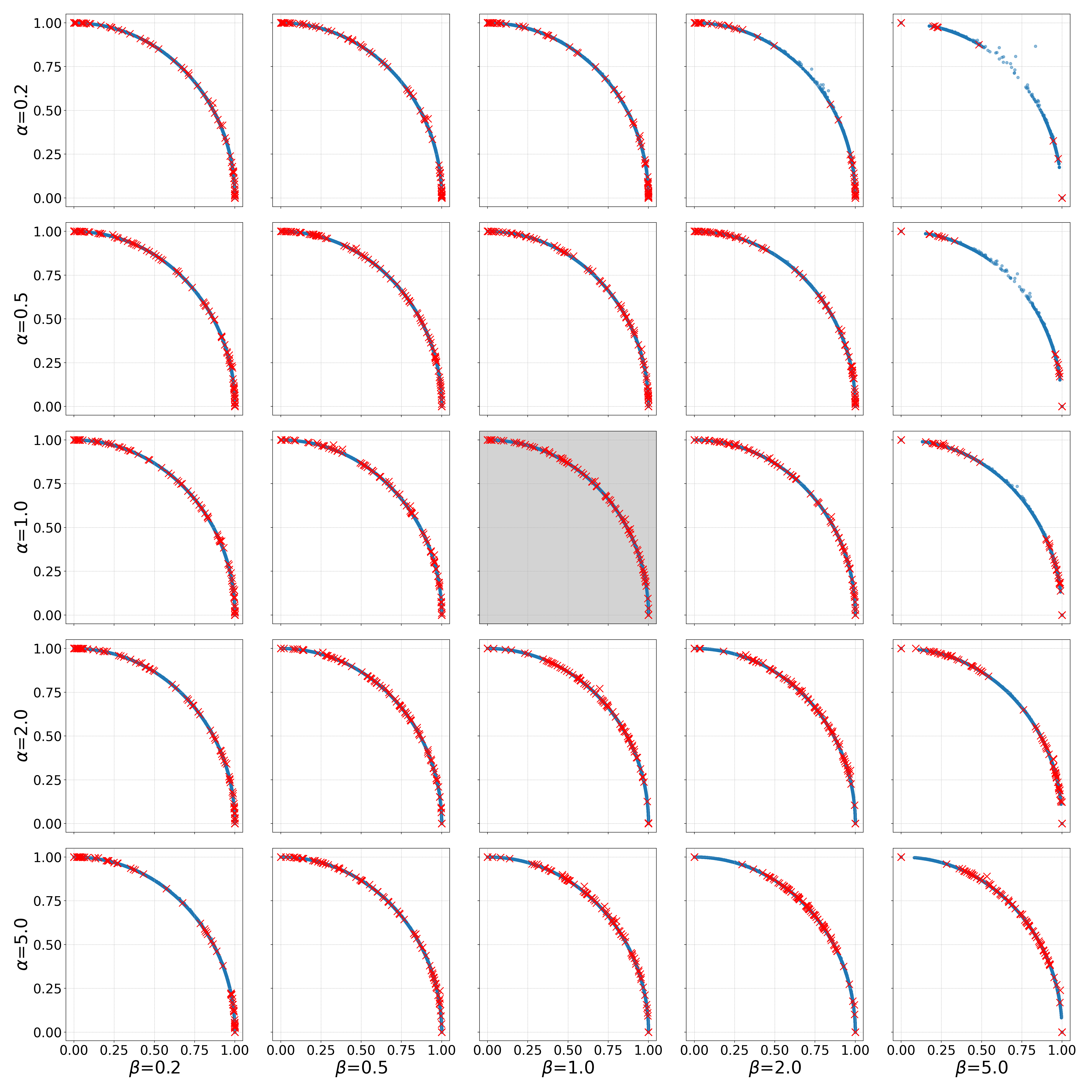}
    \caption{Objective space results from running NSGA2 (population size 10) on different objective space transformations of DTLZ2. Blue points represent non-dominated points in the unbounded archive, while red crosses represent points in the final population. Each plot corresponds to one parameterization of the BetaCDF, with the middle plot matching the default version of the base problem. Each plot contains 10 independent repetitions and shows the original objective space (not the space as seen by the algorithm). }
    \label{fig:final_points_NGSA2_example}
\end{figure}
In Figure~\ref{fig:final_hv_ab}, we show the heatmaps of the average hypervolume over 10 repetitions for each of the used algorithms on the different instantiations of the BetaCDF-based search space transformation. The base problem for this figure is the 2-dimensional ZDT3, shown as the center cell in each plot. From this plot, we can see the clear impact of this transformation on the results of random search, as when $\alpha$ is small, the achieved hypervolume decreases drastically. However, for the other algorithms, most of the search space changes have a relatively minor impact. The only exceptions for this base function are when both $\alpha$ and $\beta$ become small, in which case the final hypervolume decreases slightly. This confirms that, as expected, the changes in performance are not caused by invariances, but rather indicate stable behavior over this type of density-based change in the search space.

\subsection{Objective space transformation}

While our focus is mainly on transformations of the search space, we can apply similar methodologies to the objective space to compare their relative impacts. In particular, we can apply the BetaCDF-based transformations to the objective space, since these functions are monotonic and thus preserve the Pareto dominance relations. Since these transformations are also invertible, we can differentiate between the objective space as seen by the algorithm and the original objective space. In our analysis, we consider the original objective space (by inverting the transformation before showing the solutions or calculating any indicator values). This way, performance comparisons are not skewed by the changes in, e.g., the size of the maximum dominated region.

\begin{figure}
    \centering
    \includegraphics[width=0.99\linewidth,trim=75 20 0 50,clip]{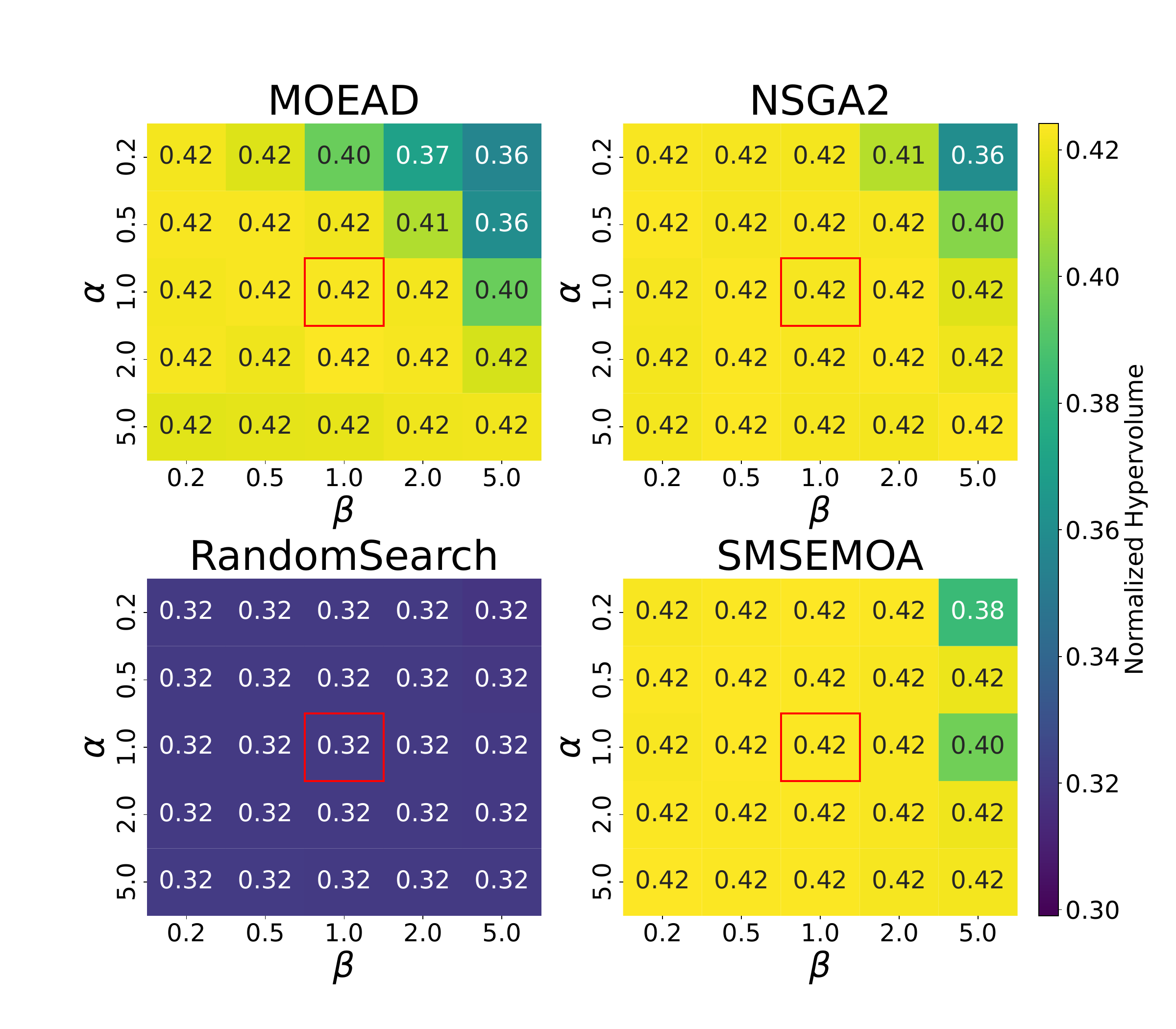}
    \caption{Final hypervolume of the archive from each of the used algorithms on different instantiations of the BetaCDF transformation of the objective space (parameterized by $\alpha$ and $\beta$ in the rows and columns, respectively). Based on the 2-dimensional DTLZ3 problem, using a population size of 10.}
    \label{fig:obj_space_hv_example}
\end{figure}
The first aspect we investigate is the extent to which the change in objective values from the algorithm's perspective changes the final solutions found. Figure~\ref{fig:final_points_NGSA2_example}, for example, shows the final Pareto front of the DTLZ2 problem and how the objective space transformations impact NSGA2's ability to find it. Notice that for the extreme cases of the transformation, the front is clearly less covered than for the original function, with the missing regions changing based on the used parameterization.

Since our analysis of the impact of this objective space transformation is based on the original rather than the transformed search space, we can also fairly compare the hypervolumes achieved by the optimizers under different instantiations of the transformation. In Figure~\ref{fig:obj_space_hv_example}, we show these hypervolumes (over the unbounded archive) on the 2-dimensional DTLZ3 problem. As expected, the random search is not impacted by these changes, while the other algorithms see some detrimental effects for more extreme versions of the transformation (making the observed Pareto front more concave from the viewpoint of the algorithms). Interesting to note are the differences between the three algorithms, with the MOEAD being the most affected. This might be related to its design, as its reference-directions based setup mean it struggles with Pareto fronts with extremely concave shapes. 

\subsection{Aggregated Results}
\begin{figure}
    \centering
    \includegraphics[width=0.95\linewidth]{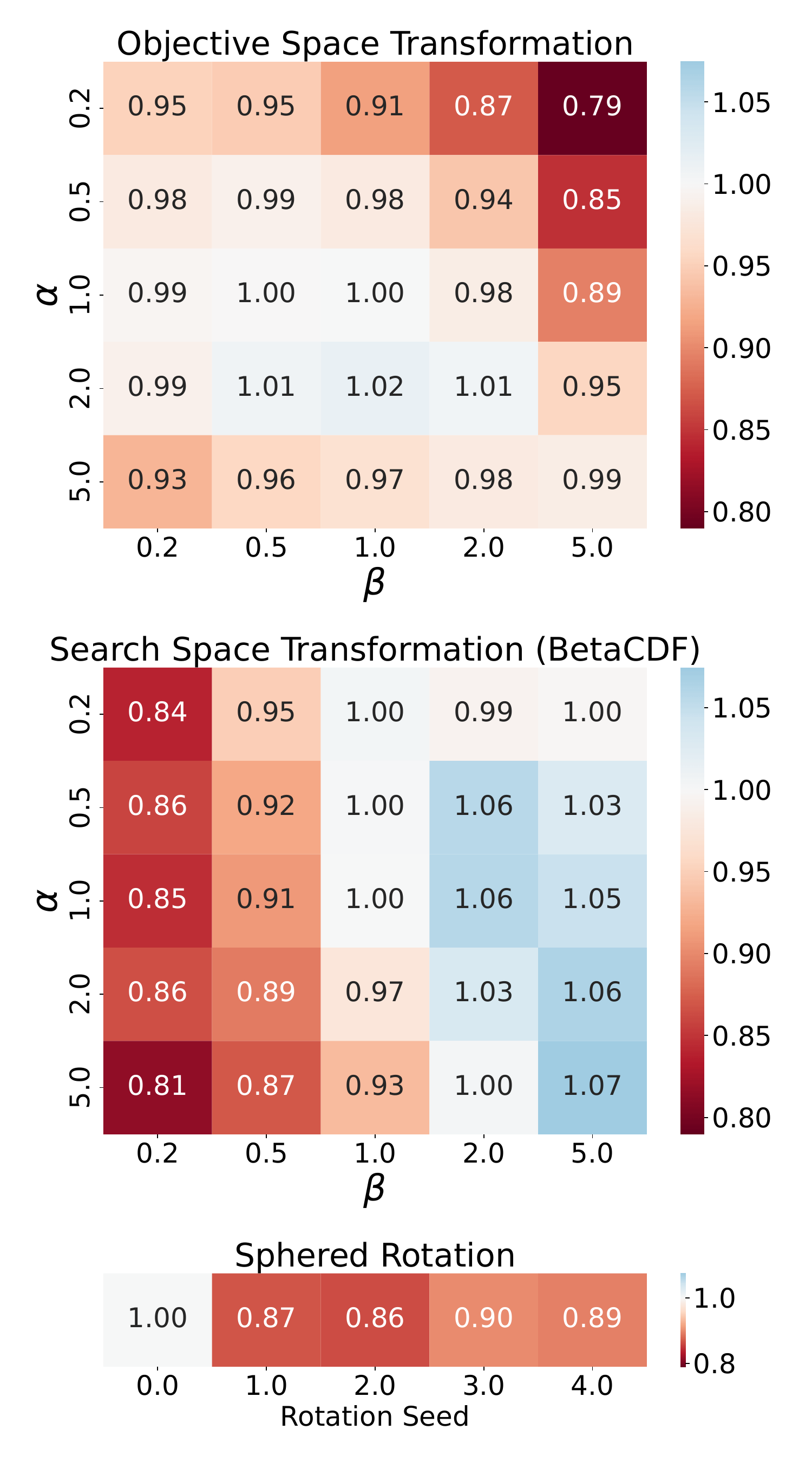}
    \caption{Average relative hypervolume (relative to the untransformed problems) for different transformations on the whole 2-dimensional DTLZ suite. Results are aggregated over the used algorithms (except random search), and use only the final population for hypervolume computation.}
    \label{fig:relative_specific}
\end{figure}
\begin{figure*}[t]
    \centering
    \includegraphics[width=0.95\linewidth]{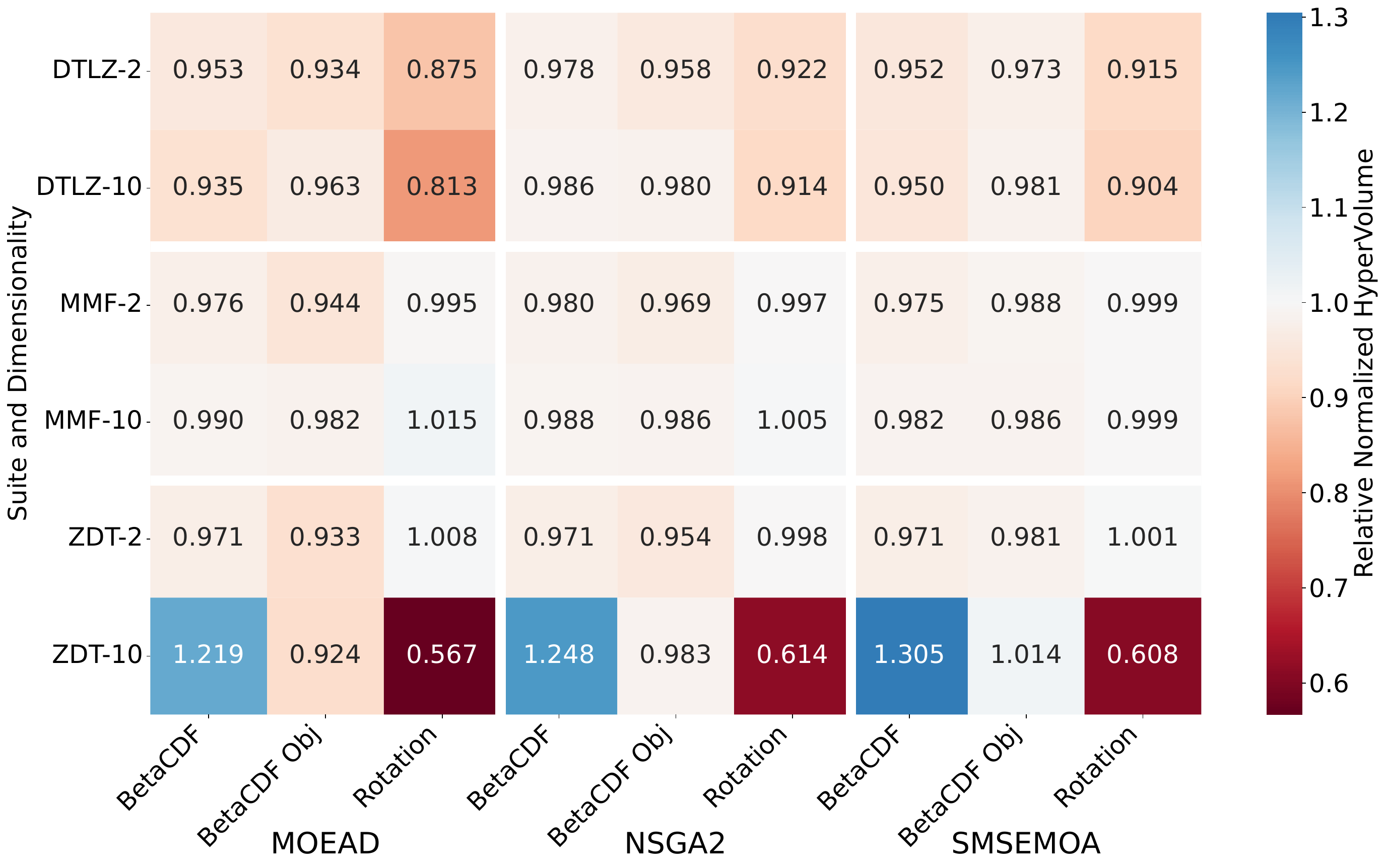}
    \caption{Averaged relative hypervolume (relative to the untransformed problems) for the 3 different transformation methods and algorithms. Aggregated over all instantiations of each transformation (equal weights for all instantiations). Hypervolume is calculated using the final population of each algorithm run. }
    \label{fig:hv_relative_agg}
\end{figure*}

In the previous sections, we analyzed the impact of each transformation type in isolation and on specific base problems. Now, we will aggregate the impacts of each transformation and identify the general trends across the used settings. Rather than looking at the absolute hypervolume, we consider the relative hypervolume achieved on transformed problems relative to the base (untransformed) function. This type of normalization allows us to aggregate over multiple problems while maintaining interpretability, as values above 1 indicate generally `easier' problems, while lower values correspond to more `difficult' problems. When we consider this aggregation over a full problem suite, such as is done in Figure~\ref{fig:relative_specific} for 2-dimensional DTLZ, we see the general trend of the different transformations.

By looking at Figure~\ref{fig:relative_specific}, we observe that changes to the objective space have a somewhat smaller impact on the reached hypervolume compared to the same magnitude of transformation of the search space. We also note that the impact of rotation does not seem to depend too much on which rotation matrix (seed) was used, although seeds 3 and 4 lead to slightly better performance. 
As can be seen in Figure~\ref{fig:rot_space_example}, these seeds correspond to rotations that are close to orthogonal, which might explain the difference. Overall, we can see that the exact rotation angle has a comparatively small impact; as long as it is not exactly orthogonal, it still makes these problems more difficult for the selected algorithms to solve. 

Finally, we can aggregate the relative hypervolume over the different parameterizations to compare the impact of the transformations on each of the problem suites. Figure~\ref{fig:hv_relative_agg} shows the mean relative hypervolume for each of these settings. From this figure, we see that the impact of transformations seems relatively stable for DTLZ when changing dimensionality, while this is not the case for the ZDT problems. This effect could be attributed to ZDT's function structure, which is highly separable. Specifically, for most ZDT problems, the first objective relies on $x_0$ and the second objective is a summation of the remaining input elements in relation to the first objective. A higher dimensionality could therefore amplify the transformation effects.

\section{Discussion}

In the previous section, we have analyzed the impact of different transformation methods on the behavior of a set of common multi-objective optimization algorithms. 
Since our transformations either impact the search or the objective space, different aspects of the considered algorithms can be investigated. 
When considering transformations on the objective space, we observed that changing the shape of the Pareto front impacts the solutions preserved until the final populations, showcasing the effects of the selection operators. Since the impact of the Pareto front shape is rather well-studied, we can confirm that this matches the observations in existing research~\cite{huband2006WFG,rookPotentialAutomatedAlgorithm2022}. 

Comparing the magnitude of changes to the objective space to the same BetaCDF transformation on the search space shows that the latter can also be very impactful. This type of analysis also highlighted the known issues of standard mutation mechanisms in having a bias towards axis-aligned steps. While previous work has shown this effect on hand-crafted rotated versions of ZDT~\cite{igel2007covariance}, the benefit of our sphered rotation transformation lies in its independence of the original problem formulation. 

By considering transformations independently from the base problem formulation, we move closer to an instance-generation mechanism for multi-objective optimization benchmarks. However, care should be taken not to translate too directly from existing mechanisms such as those in single-objective BBOB~\cite{hansen2021coco}, given the bound-constrained nature of the traditional multi-objective benchmark problems. As such, we should not expect performance to be completely invariant with respect to e.g. the sphered rotation. Rather, instance transformations such as these could be used diagnostically, aiming for robust rather than invariant performance across some set of instantiations. 

\section{Conclusions and Future Work}

In this paper, we have explored how commonly used multi-objective optimization algorithms can be impacted by changes to the search space, even when they preserve the structure of the objective space. Of particular interest is the performance deterioration observed as an effect of the sphered rotation. This suggests that there might have been implicit incentives in the used benchmark suite to design biased variation operators. While this might be useful if these same biases occur in the types of problems these algorithms aim to solve, it might be detrimental to others, and this should be taken into consideration when analyzing their behavior. 

More generally, our results highlight the need for a critical look at benchmark design and practices. While it is natural to focus multi-objective benchmark suites around the structure in the objective space, this should not prevent us from carefully designing the objective space to avoid unintended biases that can be exploited by optimization algorithms. One aspect to consider is instance generation, where rather than relying on a single problem definition, we should benchmark on transformed variations of the base problem. The question which types and magnitudes of these transformation are suitable for this purpose remains open, especially given that we might not be able to translate the invariance-perspective directly to the multi-objective setting. 

The transformations used in this paper also highlight the potential of detailed algorithmic analysis by separately changing the search and objective space. This type of analysis might allow for some degree of decomposition of problem difficulties in multi-objective optimization between the search and objective space structures. Future work should look at combinations of different transformations to identify the extent to which their interactions impact optimization behavior. 

Finally, some of the impacts of problem transformations we identified in this paper have been known about in the community for a long time. Specifically, the impact of the axis-aligned biases in variation operators has been explored almost 20 years ago~\cite{igel2007covariance}. 
This shows the need for benchmarks to evolve, and for algorithm design to more directly consider whether certain design decisions are guided by intended behavior or by benchmark performance alone~\cite{volzshort}. While important steps in this direction are being taken, e.g. through the design of new algorithmic mechanisms~\cite{doerrTheoreticalAnalysesMultiObjective,carvalhoImprovingNSGAIIAdaptive2009}, their usage has so far been limited compared to the long-established alternatives.

\begin{acks}
Diederick Vermetten is supported by funding by the European Union (ERC, ``dynaBBO'', grant no.~101125586). The authors thank the Paderborn Center for Parallel Computing (PC2) for providing computing time.
\end{acks}

\bibliographystyle{ACM-Reference-Format}
\bibliography{bibliography} 
\end{document}